\documentclass[]{bytedance_seed}

\usepackage[toc,page,header]{appendix}

\usepackage{float}
\usepackage{minitoc}
\usepackage{xspace}

\usepackage{listings}
\usepackage{xcolor}
\newcommand{\veomni}{VeOmni\xspace}

\newcommand\eg{{\it e.g.}}
\newcommand\ie{{\it i.e.}}

\title{\veomni: Scaling Any Modality Model Training with Model-Centric Distributed Recipe Zoo}

\author[*,\dagger]{Qianli Ma}
\author[*]{Yaowei Zheng}
\author[*]{Zhelun Shi}
\author[*]{Zhongkai Zhao}
\author[*]{Bin Jia}
\author[*]{Ziyue Huang}
\author{Zhiqi Lin}
\author{Youjie Li}
\author{Jiacheng Yang}
\author[\dagger]{Yanghua Peng}
\author[\dagger]{Zhi Zhang}
\author[\dagger]{Xin Liu}

\affiliation{ByteDance Seed}

\contribution[*]{Equal Contribution}
\contribution[\dagger]{Corresponding authors}

\abstract{
Recent advances in large language models (LLMs) have driven impressive progress in omni-modal understanding and generation. However, training omni-modal LLMs remains a significant challenge due to the heterogeneous model architectures required to process diverse modalities, necessitating sophisticated system design for efficient large-scale training. Existing frameworks typically entangle model definition with parallel logic, incurring limited scalability and substantial engineering overhead for end-to-end omni-modal training. 
We present \veomni, a modular and efficient training framework to accelerate the development of omni-modal LLMs. \veomni introduces model-centric distributed recipes that decouples communication from computation, enabling efficient 3D parallelism on omni-modal LLMs. \veomni also features a flexible configuration interface supporting seamless integration of new modalities with minimal code change. 
Using \veomni, a omni-modal mixture-of-experts (MoE) model with 30B parameters can be trained with over 2,800 tokens/sec/GPU throughput and scale to 160K context lengths via 3D parallelism on 128 GPUs, showcasing its superior efficiency and scalability for training large omni-modal LLMs.
}

\date{August 4, 2025}
\correspondence{\email{maqianli.fazzie@bytedance.com}}

\checkdata[Project Page]{\url{https://github.com/ByteDance-Seed/VeOmni}}

\begin{document}
\maketitle

\section{Introduction}

\begin{figure*}[t]
    \centering
    \includegraphics[width=1.0\linewidth]{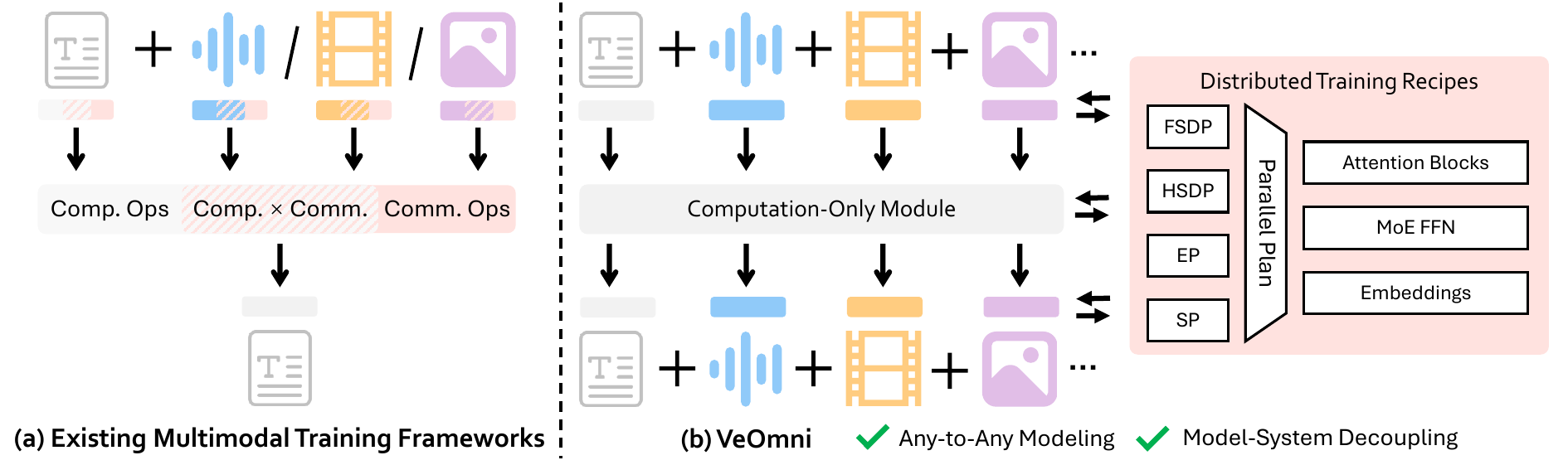}
    \caption{\textbf{Comparison between \veomni and existing training frameworks.}}
    \label{fig:Motivation}
\end{figure*}

The evolution of large language models (LLMs) has progressed from unimodal specialization towards unified omni-modal understanding and generation~\cite{sun2024emu,liu2025world,wu2025janus,xie2025show}. Recent models such as GPT-4o~\cite{hurst2024gpt4o} and BAGEL~\cite{deng2025bagel} exhibit strong performance across diverse multimodal tasks. These tasks include visual question answering~\cite{alayrac2022flamingo,guo2023imagevqa}, controllable image generation~\cite{zhang2023controllable,li2025controlar}, and multimodal reasoning~\cite{wang2024exploring,su2025thinking,huang2025vision}, highlighting the growing potential of LLMs as general-purpose omni-modal agents.

As LLMs are extended to handle diverse modalities, their architectures have become increasingly heterogeneous and complicated. State-of-the-art omni-modal LLMs typically incorporate multiple modality-specific pre-trained networks to process inputs with fundamentally different properties, such as continuous signals (\eg, images, audio) and discrete sequences (\eg, text). In these omni-modal models, a language model often serves as the central backbone~\cite{vaswani2017attention,brown2020language}, connecting with vision encoders~\cite{dosovitskiy2021image}, audio encoders~\cite{dong2018speech}, and image generation networks~\cite{ho2020denoising,goodfellow2020generative} through learned bridging mechanisms.

Despite these advances, the development of omni-modal LLMs still lags behind their unimodal counterparts. This gap is largely attributed to the absence of scalable training frameworks tailored for omni-modal tasks. While numerous mature and widely adopted systems exist for training language models on text-to-text tasks~\cite{shoeybi2019megatron,narayanan2019pipedream,narayanan2021megatron,li2023colossal,shen2024nemo,liang2025torchtitan,vescale}, few frameworks are designed to support any-to-text tasks~\cite{huang2024distmm,zhang2024disttrain,ji2024align,feng2025optimus}. Crucially, none of these frameworks enable end-to-end training for fully omni-modal scenarios, \ie, any-to-any learning, revealing the necessity for scalable infrastructure to build the next generation of omni-modal LLMs.

Extending existing training frameworks to omni-modal LLMs is non-trivial. To address the growing parameter sizes, modern training frameworks leverage various distributed training strategies~\cite{shoeybi2019megatron,narayanan2019pipedream,li2020pytorch,narayanan2021megatron} to alleviate computation and memory bottlenecks. For example, Megatron-LM~\cite{shoeybi2019megatron,narayanan2021megatron} provides optimized transformer blocks~\cite{vaswani2017attention} with advanced parallelization strategies like tensor parallelism (TP) and pipeline parallelism (PP). However, as omni-modal architectures become complex in both functionality and parameter size, directly applying these techniques to omni-modal LLMs often leads to load imbalance and poor scalability~\cite{feng2025optimus}, due to the current frameworks tightly couple model definition with parallel logic. This entanglement hiders generalization to more diverse model architectures. Although recent efforts~\cite{huang2024distmm,zhang2024disttrain,feng2025optimus} attempt to mitigate the inefficiencies of omni-modal training, they still suffer from the coupling of communication and computation, resulting in significant engineering cost and limited extensibility.

To efficiently train large omni-modal LLMs, \veomni introduces model-centric distributed recipes that decouple model definition from parallel logic. As shown in Figure~\ref{fig:Motivation}, distributed strategies such as fully sharded data parallel (FSDP), hybrid sharded data parallel (HSDP)~\cite{zhao2023pytorch,liang2025torchtitan}, sequence parallelism (SP)~\cite{jacobs2023deepspeed}, and expert parallelism (EP)~\cite{zhang2025cometfinegrainedcomputationcommunicationoverlapping} can be applied to model blocks via a high-level {\textit{parallel plan}} API. This design effectively separates communication from computation and has been validated in recent LLM training systems~\cite{liang2025torchtitan}. \veomni supports flexible composition of parallel strategies (\eg, FSDP+SP for 2D and FSDP+SP+EP for 3D parallelism), enabling a range of training recipes tailored to dense or MoE models. By decoupling parallel logic, new modality-specific modules can be integrated with minimal engineering effort, as computation modules need not account for distributed concerns. In addition, \veomni provides a light-weight interface for customizing omni-modal LLMs, allowing users to easily modify the modality-specific encoders and decoders.

Our contributions are as follows:

\begin{enumerate}
    \setlength{\itemsep}{5pt} 
    \item We propose model-centric distributed recipes that efficiently scale omni-modal training using composable n-D parallelism with minimal engineering overhead.
    \item We introduce a light-weight configuration interface for flexibly customization of omni-modal LLMs, supporting both multi-modal understanding and generation.
    \item We demonstrate \veomni's competitive efficiency and scalability across 8–128 GPUs on models ranging from 7B to 72B parameters under omni-modal scenarios.
\end{enumerate}

\clearpage

\section{Related Work}

\subsection{Multi-Modal and Omni-Modal LLMs}

Recent advances in large language models (LLMs) have increasingly focused on developing multi-modal and omni-modal capabilities. The prevailing approaches include incorporating modality-specific encoders into the input space of the LLMs for multi-modal understanding~\cite{llava,Qwen2-VL,lamm,video-llava,girdhar2023imagebind}, and attaching generative decoders to the output space for multi-modal generation~\cite{huang2024smartedit,mige} or embodied action prediction~\cite{pi_0,openvla}. More recently, omni-modal LLMs that support unified understanding and generation across modalities have emerged, aiming to align arbitrary modality features with language in a shared latent space. Based on a unified paradigm of auto-regressive modeling over multi-modal embeddings, these models differ significantly in how the modality features are encoded and decoded. Some~\cite{Chameleon_Team_Chameleon_Mixed-Modal_Early-Fusion_2024, wu2025janus} adopt discrete-token generation pipelines based on VQ-VAE series~\cite{vqgan1, razavi2019generating}. Some~\cite{wang2024illume, huang2025illume+,sun2024generative} employ continuous-token generation via latent diffusion models~\cite{dit1, dit2}. Others~\cite{bagel, mmada, transfusion} explore hybrid architectures that combine diffusion-based generation with auto-regressive decoding. In addition, some works proposed alternative next-token prediction schemes, such as masked generation~\cite{mar}, next patch prediction~\cite{patch-sd}, next scale prediction~\cite{var}, next block prediction~\cite{block-sd}, offering increased flexibility beyond standard auto-regressive decoding. Given this diverse landscape of model architectures, \veomni offers a flexible and extensible framework for building and scaling new modeling paradigms.

\subsection{LLM Training Frameworks}

The landscape of large language model training frameworks has evolved significantly to address the computational demands of scaling. For pure text training, several mature frameworks have established themselves as industry standards. Megatron-LM~\cite{shoeybi2019megatron,narayanan2021megatron} pioneered optimized transformer blocks with advanced parallelization strategies including tensor parallelism (TP) and pipeline parallelism (PP), becoming the foundation for many subsequent frameworks. Colossal-AI~\cite{li2023colossal} extends this paradigm by offering comprehensive 3D parallel training strategies that combine data, tensor, and pipeline parallelism for enhanced scalability. NeMo~\cite{shen2024nemo} provides an end-to-end cloud solution specifically designed for training large-scale LLMs with enterprise-grade features, while TorchTitan~\cite{liang2025torchtitan} and veScale~\cite{vescale} represent the newer generation of PyTorch-native frameworks that emphasize auto-parallelization and simplified programming models. 

In the multi-modal domain, specialized frameworks have emerged to handle the unique challenges of any-to-text training. DistMM~\cite{huang2024distmm} introduces optimizations specifically tailored for multimodal LLM training, while DistTrain~\cite{zhang2024disttrain} addresses model and data heterogeneity through disaggregated training approaches. Align Anything~\cite{ji2024align} focuses on cross-modality model training with feedback mechanisms, and Optimus~\cite{feng2025optimus} accelerates large-scale multi-modal LLM training through bubble exploitation techniques. 
However, these existing frameworks primarily target either text-to-text or any-to-text scenarios, leaving a significant gap in supporting comprehensive omni-modal training (any-to-any) scenarios. Our framework addresses this limitation by providing a scalable and modular solution specifically designed for the complexities of omni-modal training, enabling seamless integration of diverse modalities within a unified training pipeline.

\section{\veomni: Scalable Omni-Modal Training Framework}

\veomni designs a model-centric framework that is natively suitable for training omni-modal models. This framework includes a diverse range of distributed training strategies, enabling any modality to be easily scaled up for large-scale clusters.

\subsection{Light-Weight Customization for Omni-Modal LLMs}

\begin{figure*}[t]
    \centering
    \includegraphics[width=0.8\linewidth]{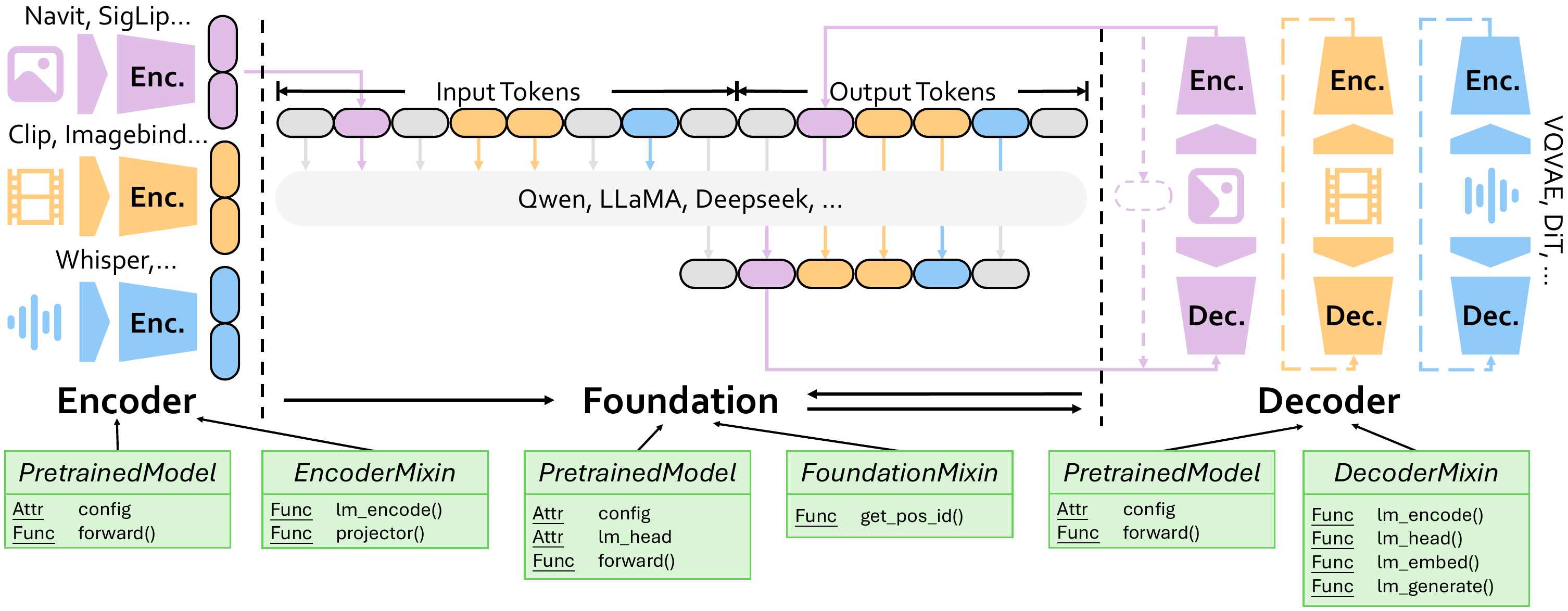}
    \caption{\textbf{Composite architecture for omni-modal LLMs}. The architecture consists of three fully decoupled modules: encoder, foundation model, and decoder.}
    \label{fig:multimodal_framework}
    \vspace{-2mm}
\end{figure*}

To support training across arbitrary modalities, we propose a plug-and-play architecture that allows any combination of multimodal encoders and decoders to be flexibly attached to the input and output sides of a foundation model, as shown in Figure~\ref{fig:multimodal_framework}. This modular and fully decoupled design decouples system-level operations from model-specific computation, allowing the streaming pipeline to operate independently of the model pipeline. Therefore, \veomni enables easy extension and scaling of diverse omni-modal LLMs.

Our core design centers around a lightweight protocol tailored for each component. Specifically, encoder and decoder modules implement a unified interface by extending the \texttt{PreTrainedModel} class from HuggingFace along with a task-specific mixin. This design ensures compatibility with standard training workflows and enables all modules to be registered, initialized, and executed consistently.

In the training phase, for input modalities, each encoder implements an \texttt{lm\_encode} function that encodes raw modality data into token embeddings, which are then inserted into the input embeddings of the foundation model. Similarly, decoder modules implement the \texttt{lm\_encode} function to provide training-time inputs representing the target output tokens. The final outputs from the foundation model are decoded into the target modality using \texttt{lm\_head} function, completing the end-to-end mapping.

In the inference phase, for each prediction step, the output hidden states is passed through the corresponding modality decoder's \texttt{lm\_embed} function, which generates embeddings that are used as inputs for the foundation model's next token prediction. Once all tokens have been predicted, the decoder's \texttt{lm\_generate} function is invoked to generate the final modality-specific output.
The details of the training and inference processes are listed in Appendix~\ref{appendix:Omni-model_train-infer}.

\subsection{Model-Centric Distributed Recipe Zoo}

To support large-scale distributed training of omni-modal LLMs, \veomni offers a modular suite of distributed training strategies, including fully sharded data parallel (FSDP), sequence parallelism (SP), and expert parallelism (EP), designed to handle challenges such as training with ultra-long sequences and efficiently scaling large mixture-of-experts (MoE) models. \veomni enables users to freely compose distributed training recipes tailored to their model architectures and modality-specific requirements, without needing to manage low-level implementation details. The framework is designed with the following principles:

\begin{enumerate}
\item \textbf{\textit{Non-intrusive distributed training APIs.}} \veomni decouples the model architecture from the underlying distributed training mechanisms, enabling users to compose omni-modal models flexibly without modifying low-level parallelization code.
\item \textbf{\textit{Support for long sequence training.}} \veomni accommodates training workloads involving extremely long sequences across different modalities, which is a common challenge in omni-modal training.
\item \textbf{\textit{Efficient scaling for MoE models.}} \veomni facilitates efficient training of large-scale MoE models through integrated expert parallelism, achieving both scalability and compute efficiency.
\end{enumerate}

In the following sections, we introduce each of these distributed technologies in \veomni and demonstrate how they are integrated to support scalable omni-modal training.

\subsubsection{Fully Sharded Data Parallel for Large Model Training}

Fully sharded data parallel (FSDP) is a highly efficient implementation of the zero redundancy optimizer (ZeRO)~\cite{rajbhandari2020zero} in PyTorch. Its primary advantage is the significant reduction in memory required on each GPU during training. By sharding a model's parameters, gradients, and optimizer states across all available devices, FSDP allows for the training of extremely large models that would otherwise exceed the memory capacity of a single GPU. A key advantage of FSDP is its non-intrusive design, which decouples the model's architecture from the parallelization strategy. This means developers can focus on designing their models without needing to modify the underlying code to accommodate the distributed training setup. This non-intrusive nature makes FSDP exceptionally well-suited for training omni-modal LLMs. Since these models are defined by their complex and heterogeneous architectures, FSDP's ability to parallelize training without requiring any changes to the model's code makes it an ideal solution. \veomni integrates both FSDP1~\cite{zhao2023pytorch} and FSDP2~\cite{liang2025torchtitan} as fundamental components of its distributed training stack, and provides a unified API for easy configuration. More details in Appendix~\ref{appendix:FSDP}.

To further improve training efficiency, \veomni integrates hybrid sharded data parallel (HSDP), an extension of FSDP designed to minimize communication overhead. HSDP utilizes a 2D device mesh, combining FSDP within ``shard groups'' and distributed data darallel (DDP) across ``replicate groups''. This hybrid approach drastically cuts down on inter-node communication, enabling even greater scalability. The switch to the more efficient HSDP is as simple as changing one parameter in the configuration,  with no modifications to the underlying code required.

\subsubsection{Sequence Parallelism for Long Sequence Training}

With the advancement of state-of-the-art omni-modal LLMs, the sequence lengths required for training have grown significantly, particularly in domains such as high-resolution image or video understanding and generation~\cite{Qwen2-VL,llava-video,gao2025seedance}. This trend toward longer sequences poses substantial challenges in terms of both computational cost and memory consumption. Efficiently addressing the demands of long sequence training is therefore critical for scaling model capacity and unleashing the full potential of omni-modal architectures. To address this, \veomni adopts DeepSpeed-Ulysses~\cite{jacobs2023deepspeed}, a highly efficient sequence parallelism technique specifically designed for transformer training on ultra-long sequences. DeepSpeed-Ulysses splits activations along the sequence dimension and strategically orchestrates all-to-all communications during attention computation, ensuring that communication volume remains constant when both sequence length and device count are scaled proportionally. This design enables high efficiency and scalability for long-sequence training. A core advantage of \veomni is its provision of a non-intrusive interface for deploying DeepSpeed-Ulysses. Developers can seamlessly leverage DeepSpeed-Ulysses without needing to modify model-building code or directly interact with the underlying implementation details. In particular, \veomni enhances the implementation of FlashAttention~\cite{dao2022flashattention} by integrating DeepSpeed-Ulysses while keeping fully compatible with the default attention, enabling the straightforward adoption of advanced attention acceleration methods within sequence parallel workflows. More details in Appendix~\ref{appendix:ulysess}.

Furthermore, to optimize training throughput, we introduce Async-Ulysses, an enhanced implementation designed to overlap communication and computation. This version schedules the all-to-all communication operations to execute concurrently with the linear projection computations. By effectively hiding a significant portion of the communication latency behind computation, this approach yields substantial improvements in training performance.

\subsubsection{Expert Parallelism for MoE Model Scaling}

Mixture-of-Experts (MoE) architectures have emerged as a mainstream solution for scaling large models efficiently~\cite{deepseekai2025deepseekv3technicalreport}, particularly due to their ability to sparsely activate subsets of parameters during inference and training, enabling significant improvements in both computational efficiency and model capacity. In the context of omni-modal foundation models, adopting MoE architectures as the backbone not only reduces training cost but also enhances model performance across diverse modalities by dynamically allocating expert capacity based on input content.

To support the scalable training of MoE-based omni-modal LLMs, \veomni introduces a user-friendly and flexible interface for expert parallelism, allowing users to easily apply expert sharding across devices without manual configuration. This interface is compatible with widely used MoE variants and supports plug-and-play integration, thereby significantly lowering the barrier to implementing distributed expert parallelism. More details in Appendix~\ref{appendix:ep-parallel-plan}.

A major bottleneck in large-scale MoE training lies in all-to-all communication required for routing tokens to their assigned experts, introducing substantial latency. To mitigate this, \veomni incorporates fine-grained communication-computation overlapping techniques inspired by recent advances~\cite{zhang2025cometfinegrainedcomputationcommunicationoverlapping,chang2024fluxfastsoftwarebasedcommunication}. These works hide communication latency by scheduling collective operations concurrently with local expert computation, eliminating the need for complex pipeline-level solutions such as DualPipe~\cite{deepseekai2025deepseekv3technicalreport}. Unlike pipeline-centric designs, which are often rigid and sensitive to modality-specific imbalance, \veomni's operator-level approach is lightweight, model-agnostic, and particularly well-suited to multi-modal settings. This results in higher utilization and faster iteration during large-scale MoE training.

\subsubsection{Composable n-D Parallel Training}

\begin{figure*}[t]
    \centering
    \includegraphics[width=0.9\linewidth]{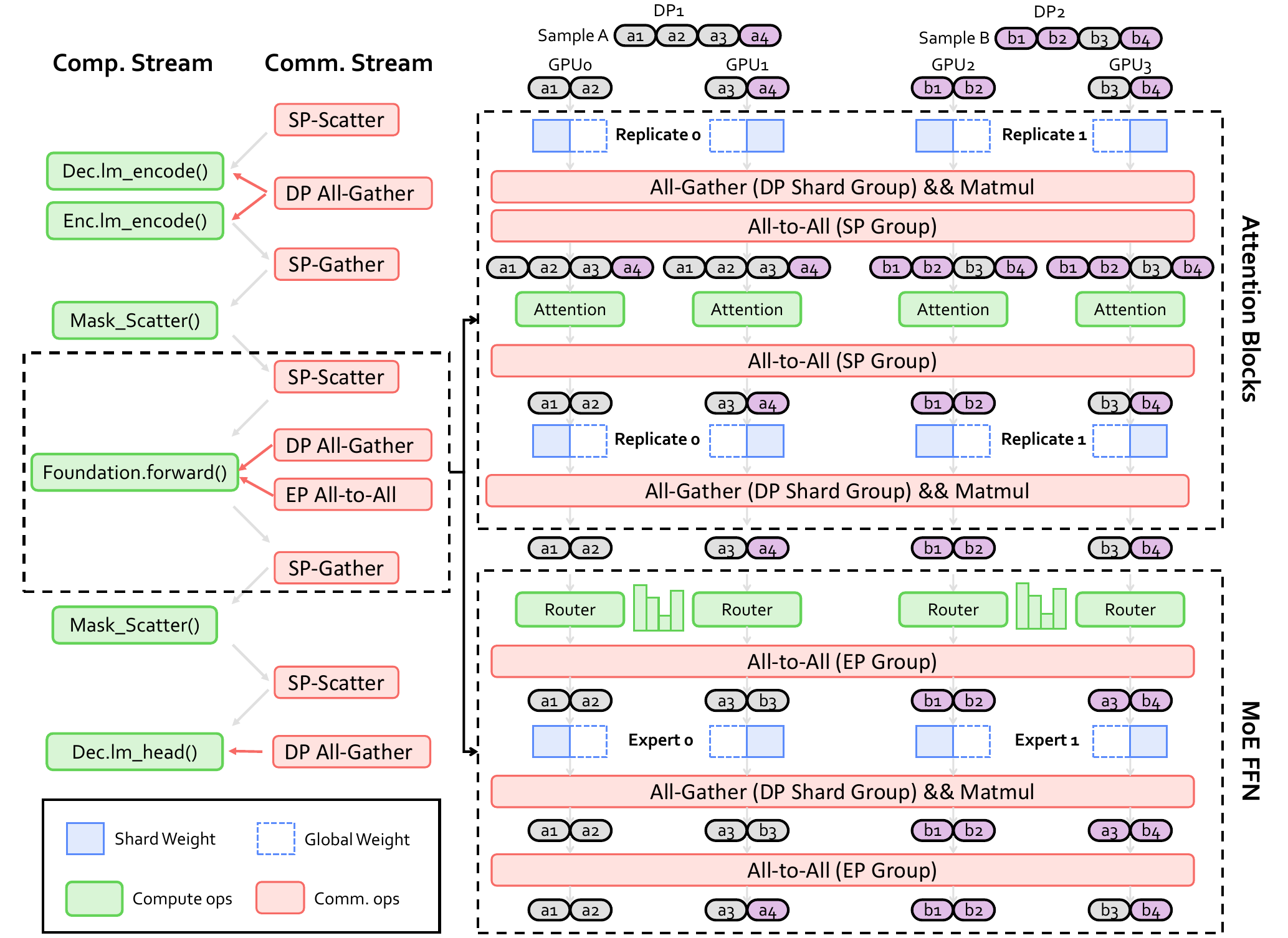}
    \caption{
        \textbf{Overview of \veomni's n-D parallelism.}
        The figure on the left illustrates the data flow of \veomni during the training of omni-modal LLMs. The encoder for each modality processes its respective inputs and then scatters the features to the corresponding ranks through an all-to-all communication operation.
        The right figure shows the 3D parallelism of \veomni, where attention blocks apply HSDP with a data sharded size of 2 and a data replicate size of 2, inputs apply sequence parallelism with a size of 2, and mixture-of-experts blocks apply expert parallelism and FSDP with an EP size of 2 and a data sharded size of 2.
    }
    \label{fig:3DParallel}
    \vspace{-2mm}
\end{figure*}

In \veomni, core parallelism strategies (\ie, fully sharded data parallel (FSDP), sequence parallelism (SP), and expert parallelism (EP)) are designed to be fully composable and can be flexibly applied to different components of an omni-modal LLM. This composability is particularly advantageous for omni-modal training, where different modalities or architectural components may benefit from distinct parallelism techniques. For example, vision encoders can adopt FSDP or standard data parallelism depending on their scale, while language backbones can leverage a combination of EP for MoE layers and SP to support long sequence processing. This fine-grained flexibility enables efficient and scalable training across diverse model architectures.

Figure~\ref{fig:3DParallel} provides an overview of \veomni's n-D parallelism design. The left side of the figure illustrates the communication and computation flow for a omni-modal model, where different modality-specific encoders process their respective inputs and distribute intermediate features to downstream modules. The right side visualizes the application of n-D parallelism strategies across different parts of the model, demonstrating how various parallelism techniques coexist within a single training configuration. To support this flexible parallelism composition, \veomni introduces a unified abstraction for distributed training based on a global device mesh~\cite{liang2025torchtitan}. Unlike approaches~\cite{shoeybi2019megatron, narayanan2021megatron} that require manually managing multiple process groups, \veomni significantly simplifies the configuration and coordination of n-D parallelism (Appendix~\ref{appendix:parallel_state}). This not only reduces the complexity of process group management but also improves extensibility, making it easier to adapt and extend to future parallelism strategies or new model components.

\subsubsection{Other System Optimization Strategies}

In addition to the aforementioned parallelism strategies, \veomni also incorporates a wide range of other system-level optimizations, following the key design principle of \veomni, which is the decoupling of these optimization implementations from the model's computation logic. This modular architecture allows for seamless integration, enabling these system-level enhancements to be readily applied across various model architectures with minimal to no modification of the model code. These optimizations include, but are not limited to, the following strategies:

\begin{itemize}
    \item \textbf{\textit{Dynamic Batching.}} Padding all samples in a batch to a fixed sequence length often leads to substantial computational inefficiency, particularly when there is significant variation in sequence lengths across samples. To mitigate this issue and improve training efficiency, \veomni employs a dynamic batching mechanism that accumulates samples in a buffer and strategically packs them to approximate a target sequence length. Leveraging FlashAttention~\cite{dao2023flashattention}, this packing strategy enables efficient utilization of the batch budget with minimal padding overhead, while preserving the correctness of attention computation across samples.
    \item \textbf{\textit{Efficient Kernels.}} To maximize training throughput, \veomni incorporates a suite of highly optimized operator kernels, including RMSNorm, LayerNorm, RoPE, SwiGLU, and in-place cross-entropy from liger-kernel~\cite{hsu2025ligerkernelefficienttriton}, along with FlashAttention~\cite{dao2022flashattention,dao2023flashattention,shah2024flashattention} and MoE-specific operations~\cite{chang2024fluxfastsoftwarebasedcommunication, zhang2025cometfinegrainedcomputationcommunicationoverlapping}. These kernels are carefully engineered for both high performance and broad compatibility, enabling efficient execution across diverse transformer-based architectures and model variants.
    \item \textbf{\textit{Memory Optimization.}} \veomni incorporates layer-wise recomputation~\cite{chen2016training}, activation offloading, and optimizer state offloading to substantially reduce memory consumption during training. These memory-saving techniques enable the use of larger micro-batch sizes per GPU, which in turn improves the amortization of communication costs and facilitates better overlap with the communication overhead introduced by fully sharded data parallel (FSDP), ultimately enhancing overall training efficiency.
    \item \textbf{\textit{Efficient Distributed Checkpointing.}} \veomni leverages ByteCheckpoint~\cite{wan2024bytecheckpoint} to enable efficient checkpoint saving and resumption across varying distributed configurations with minimal overhead. Beyond facilitating elastic training and enhancing cluster utilization, our framework further extends ByteCheckpoint to support omni-modal models. This extension ensures consistent and reliable saving and loading of heterogeneous model components.
    \item \textbf{\textit{Meta Device Initialization.}} \veomni supports model initialization on the meta device for large models without allocating physical memory during initialization. After instantiating the model on the meta device, we perform parameter sharding and parallel loading by converting parameters into the DTensor format~\cite{liang2025torchtitan}, significantly accelerating the initialization and loading processes for large-scale models.
\end{itemize}

\clearpage

\section{Experiments}

In this section, we present experimental results highlighting the performance and scalability of \veomni.

\subsection{Experimental Setup}

\textbf{Environments.} We evaluate the training performance and scalability of \veomni on large-scale productive GPU clusters across configurations ranging from 8 to 128 GPUs, enabling a comprehensive study of \veomni's behavior under both moderate and large-scale distributed settings.

\textbf{Models and Datasets.} We evaluate a diverse set of model architectures, including dense models such as Qwen2-VL 7B and 72B~\cite{Qwen2-VL}, and a mixture-of-experts (MoE) omni-modal LLM based on Qwen3-30B-A3B~\cite{qwen3technicalreport}. To assess the multimodal capabilities of \veomni, we use the following domain specific datasets: FineWeb-100T~\cite{fineweb100t} for text understanding, ShareGPT4V~\cite{chen2024sharegpt4v} for image understanding, LLaVA-Video~\cite{llava-video} for video understanding, Voice Assistant~\cite{voiceassistant} for audio understanding, and ImageNet~\cite{deng2009imagenet} for image generation tasks.
Each model is adapted using its native instruction template and augmented with special tokens to delineate modality boundaries (\eg ~\texttt{<image\_start>} and \texttt{<image\_end>}).

\textbf{Workloads and Metrics.} To assess scalability across different model sizes and GPU counts, we progressively scale the input context length from 8K to 256K tokens. We evaluate the performance and scalability of \veomni by measuring training throughput (tokens per second per GPU) and model FLOPs utilization (MFU)~\cite{chowdhery2023palm}, which together provide a comprehensive view of system efficiency under varying workloads and parallel strategies. Additionally, we analyze loss convergence behavior across three structurally distinct omni-modal LLMs to validate training stability and overall effectiveness.
For all experiments, we freeze the modality-specific encoders and decoders, while fully fine-tuning the remaining components, including the foundation model and multimodal projectors.

\subsection{Comparison of Training Recipes under Different Scenarios}

\begin{figure*}[t]
    \centering
    \includegraphics[width=0.8\linewidth]{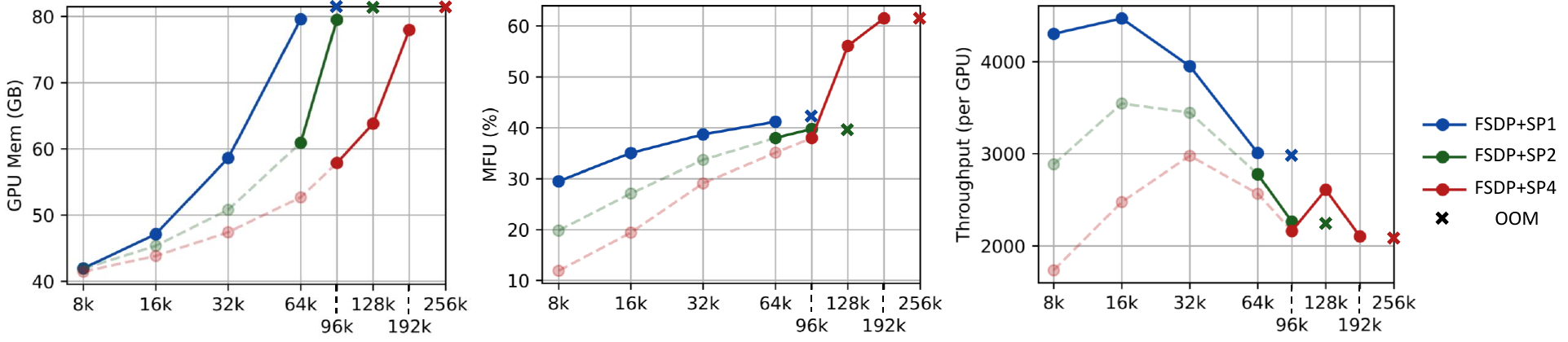}
    \caption{\textbf{2D Parallelism (FSDP+SP) with Qwen2-VL~\cite{Qwen2-VL} 7B on 8 GPUs.}
    The maximum context length varies from 8K to 256K, with allowed sequence parallel sizes ranging from 1 to 4.}
    \label{fig:qwen2vl_7b_8gpu_exp}
\end{figure*}

\begin{figure*}[t]
    \centering
    \includegraphics[width=0.8\linewidth]{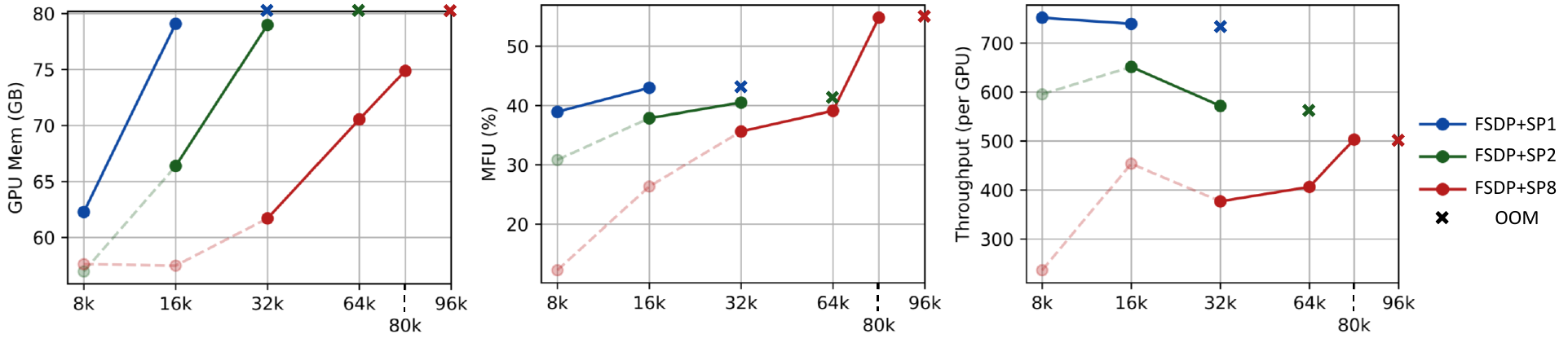}
    \caption{\textbf{2D Parallelism (FSDP+SP) with Qwen2-VL~\cite{Qwen2-VL} 72B on 128 GPUs.}
    The maximum context length varies from 8K to 96K, with allowed sequence parallel sizes ranging from 1 to 8.}
    \label{fig:qwen2vl_72b_128gpu_exp}
\end{figure*}

Figures~\ref{fig:qwen2vl_7b_8gpu_exp}–\ref{fig:multimodal_qwen3moe_30b_128gpu_exp} (a) present a comprehensive comparison of the efficiency of various training recipes across different model sizes and GPU numbers. Figures~\ref{fig:qwen2vl_7b_8gpu_exp}-\ref{fig:qwen2vl_72b_128gpu_exp} show results of training Qwen2-VL~\cite{Qwen2-VL} on 8 and 128 GPUs, respectively. The findings reveal that increasing the degree of sequence parallelism enables the models to handle significantly longer context lengths. In particular, \veomni can support up to 192K context length for training the 7B model with an MFU of 61.5\% and 96K context length for training the 72B model with an MFU of 54.82\%. Figure~\ref{fig:multimodal_qwen3moe_30b_128gpu_exp} further extends this analysis to a 3D parallelism configuration (FSDP+SP+EP) using a 30B parameter a mixture-of-experts (MoE) omni-modal LLM based on Qwen3-30B-A3B~\cite{qwen3technicalreport}, on 128 GPUs. In this setting, combinations involving moderate levels of SP and EP achieve a balanced trade-off, supporting extended context lengths of up to 160K, while maintaining competitive throughput.

The experimental results underscore the efficiency of \veomni in handling long-sequence training and scaling mixture-of-experts (MoE) models. Specifically, \veomni demonstrates strong performance by minimizing overhead in short sequence scenarios, while effectively leveraging sequence and expert parallelism to maintain scalability and feasibility in long sequence and MoE settings.

\subsection{Convergence Study on Omni-Modal LLMs}

\begin{figure*}[t]
    \centering
    \includegraphics[width=.9\linewidth]{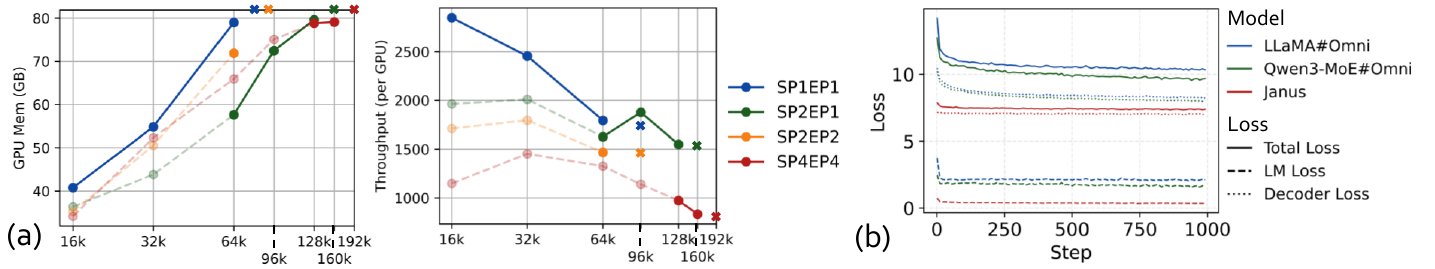}
    \caption{(a) 3D parallelism (FSDP+SP+EP) on a 30B MoE-based omni-modal LLM with 128 GPUs. The maximum sequence length varies from 16K to 192K, with allowed sequence parallel sizes ranging from 1 to 4 and expert parallel sizes ranging from 1 to 4. (b) Convergence study on three distinct omni-modal LLMs.}
    \label{fig:multimodal_qwen3moe_30b_128gpu_exp}
    \vspace{-2mm}
\end{figure*}

Figure~\ref{fig:multimodal_qwen3moe_30b_128gpu_exp} (b) presents the convergence behavior of three omni-modal LLMs on multimodal understanding and generation tasks. The understanding tasks span four modalities, \ie, text, image, video, and audio, while the generation tasks involve text and image synthesis. Janus~\cite{wu2025janus} is trained solely on image understanding and image generation. LLaMA\#Omni and Qwen3-MoE\#Omni are two custom models that share similar architectures: Qwen2.5-Omni~\cite{Qwen25Omni} ViT serves as the image and video encoder, Qwen2.5-Omni Whisper as the audio encoder, and MoVQGAN~\cite{zheng2022movq} as the image decoder. LLaMA\#Omni uses LLaMA~\cite{touvron2023llama} as the foundation model, while Qwen3-MoE\#Omni adopts Qwen3-30B-A3B~\cite{qwen3technicalreport}. Janus adopts SigLip~\cite{zhai2023sigmoid} as the image encoder, LLaMA as the foundation model, and LlamaGen~\cite{sun2024autoregressive} as the image decoder. ``LM loss'' refers to the cross-entropy loss over text tokens, and ``decoder loss'' refers to the cross-entropy loss over image tokens. As shown in Figure~\ref{fig:multimodal_qwen3moe_30b_128gpu_exp} (b), all models exhibit stable convergence across both understanding and generation tasks, demonstrating that \veomni enables efficient and robust training for large omni-modal LLMs.





\section{Conclusion}

In this work, we introduce \veomni, a model-centric distributed training framework designed to scale any-modality models efficiently. By integrating multiple parallelism methods and system optimizations into a composable recipe-based design, \veomni enables seamless and efficient distributed training strategies across omni-modal LLMs. 
We further demonstrate the system-level optimizations, modular APIs, and real-world training applications on large-scale vision-language-audio models. Experimental analysis shows that \veomni not only achieves high throughput and scalability but also provides developer-friendly abstractions for fast prototyping and production-scale deployment.
In future work, we plan to extend \veomni to support non-intrusive pipeline parallelism, enabling further decoupling of model definition and parallel execution. Additionally, we aim to enhance sequence parallelism with modality-aware data balancing strategies to better support multimodal training scenarios.

\section*{Acknowledgments}
We thank Yawei Wen, Junda Feng, Junda Zhang, Haotian Zhou, Huiyao Shu, Hongyu Zhu, Changyue Liao, Yuanhang Gao, Weidong Chen, Shu Yun, Jie Wang, Yifan Pi, Ziyi Wang, Baixi Sun as well as other colleagues at ByteDance for their contribution for the VeOmni project.

\clearpage

\bibliographystyle{plainnat}
\bibliography{main}

\clearpage

\beginappendix

\appendix

\section{Comparison of \veomni and TorchTitan on Training Large Language Models}
\label{appendix:A}

In the main paper, we demonstrate that \veomni supports efficient large-scale training for omni-modal models, a capability not yet supported by other existing frameworks. 
As such, a direct comparison in omni-modal training scenarios is infeasible. 
To provide a fair benchmark, we conduct controlled experiments on large-scale text-only models to compare \veomni with TorchTitan~\cite{liang2025torchtitan}, a state-of-the-art distributed training framework.

\begin{table}[H]
\centering
\caption{\textbf{Performance comparison between \veomni and TorchTitan with Qwen2-7B~\cite{Qwen2-VL} on 128GPUs.} Both configurations adopt mixed-precision training and full activation checkpointing, with a micro-batch size of 1 and a global batch size of 128.}
\label{tab: qwen2-7b table}
\begin{tabular}{lrrrrrrr}
\toprule
\multirow{2}{*}{\textbf{Method}}  & \multirow{2}{*}{\textbf{Seqlen}}  & \multicolumn{2}{c}{\textbf{Memory (GB)}} & \multicolumn{2}{c}{\textbf{Throughput}} & \multicolumn{2}{c}{\textbf{MFU (\%)}} \\
         &      & \textbf{\veomni} & \textbf{TorchTitan}  & \textbf{\veomni} & \textbf{TorchTitan} & \textbf{\veomni} & \textbf{TorchTitan}   \\
\midrule 
FSDP      & 8k         &  21.91          &  23.26     & \textbf{6,940}                        & 6,639       &  38.76      & 34.61  \\
FSDP+SP4  & 8k         &  23.23          &  22.55     & \textbf{3,830}                        & 3,590       &  20.66      & 18.72  \\
FSDP      & 16k        &  24.45          &  39.58     & \textbf{6,583}                        & 6,082       &  43.22      & 37.62  \\
FSDP+SP4  & 16k        &  23.92          &  25.08     & \textbf{4,889}                        & 4,056       &  31.92      & 25.24  \\
FSDP      & 32k        &  31.41          &  71.81     & \textbf{5,315}                        & 4,878       &  44.42      & 39.68  \\
FSDP+SP4  & 32k        &  25.27          &  30.13     & \textbf{4,629}                        & 3,727       &  38.69      & 30.32  \\
FSDP      & 64k        &  43.00          &  OOM       & \textbf{3,725}                        & -        &  45.92      & -    \\
FSDP+SP4  & 64k        &  28.70          &  44.82     & \textbf{3,452}                        & 2,853       &  42.35      & 34.32  \\
FSDP      & 128k       &  70.26          &  OOM       & \textbf{2,258}                        & -        &  44.95      & -    \\
FSDP+SP4  & 128k       &  35.49          &  OOM       & \textbf{2,187}                        & -        &  43.49      & -    \\
\bottomrule
\end{tabular}
\end{table}

\begin{table}[H]
\centering
\caption{\textbf{Performance comparison between \veomni and TorchTitan with Qwen2.5-32B~\cite{qwen2025qwen25technicalreport} on 128GPUs.} Both configurations adopt mixed-precision training and full activation checkpointing, with a micro-batch size of 1 and a global batch size of 128.}
\label{tab: qwen25-32b table}
\begin{tabular}{lrrrrrrr}
\toprule
\multirow{2}{*}{\textbf{Method}}  & \multirow{2}{*}{\textbf{Seqlen}}  & \multicolumn{2}{c}{\textbf{Memory (GB)}} & \multicolumn{2}{c}{\textbf{Throughput}} & \multicolumn{2}{c}{\textbf{MFU (\%)}} \\
         &      & \textbf{\veomni} & \textbf{TorchTitan}  & \textbf{\veomni} & \textbf{TorchTitan} & \textbf{\veomni} & \textbf{TorchTitan}   \\
\midrule 
FSDP+SP4 & 8k         & 35.98  & 34.37       & \textbf{1,140}   &  1,017     &  26.05   & 22.90\\
FSDP+SP8 & 8k         & 36.47  & 33.42       & \textbf{668}    &  596      &  15.14   & 13.43\\
FSDP+SP4 & 16k        & 37.43  & 39.48       & \textbf{1,316}   &  1,131     &  34.52   & 29.05\\
FSDP+SP8 & 16k        & 37.31  & 37.14       & \textbf{1,074}   &  789      &  27.87   & 20.28\\
FSDP+SP4 & 32k        & 42.42  & 46.78       & \textbf{1,240}   &  1,022     &  36.11   & 32.75\\
FSDP+SP8 & 32k        & 38.76  & 44.04       & \textbf{1,115}   &  821      &  40.36   & 26.31\\
FSDP+SP4 & 64k        & 49.06  & 68.16       & \textbf{957}    &  817      &  43.02   & 36.58\\
FSDP+SP8 & 64k        & 43.75  & 58.58       & \textbf{903}    &  689      &  40.74   & 30.84\\
FSDP+SP4 & 128k       & 64.85  & OOM         & \textbf{632}    &  -      &  44.20   & -  \\
FSDP+SP8 & 128k       & 50.41  & 79.64       & \textbf{613}    &  507      &  42.92   & 35.57\\
\bottomrule
\end{tabular}
\end{table}

\begin{table}[H]
\centering
\caption{\textbf{Performance comparison between \veomni and TorchTitan with Qwen2-72B~\cite{xu2025qwen2} on 128GPUs.} Both configurations adopt mixed-precision training and full activation checkpointing, with a micro-batch size of 1 and a global batch size of 128.}
\label{tab: qwen25-72b table}
\begin{tabular}{lrrrrrrr}
\toprule
\multirow{2}{*}{\textbf{Method}}  & \multirow{2}{*}{\textbf{Seqlen}}  & \multicolumn{2}{c}{\textbf{Memory (GB)}} & \multicolumn{2}{c}{\textbf{Throughput}} & \multicolumn{2}{c}{\textbf{MFU (\%)}} \\
         &      & \textbf{\veomni} & \textbf{TorchTitan}  & \textbf{\veomni} & \textbf{TorchTitan} & \textbf{\veomni} & \textbf{TorchTitan}   \\
\midrule
FSDP+SP4 & 8k   & 58.62  & 54.08   & \textbf{565}   & 499  &  28.04        & 24.74\\
FSDP+SP8 & 8k   & 58.35  & 51.73   & \textbf{330}   & 297  &  16.11        & 14.75\\
FSDP+SP4 & 16k  & 62.15  & 64.35   & \textbf{624}   & 544  &  35.34        & 30.44\\
FSDP+SP8 & 16k  & 59.98  & 58.02   & \textbf{532}   & 394  &  29.82        & 22.03\\
FSDP+SP4 & 32k  & 68.35  & 79.36   & \textbf{592}   & 497  &  41.05        & 34.16\\
FSDP+SP8 & 32k  & 63.49  & 70.39   & \textbf{533}   & 398  &  36.72        & 27.31\\
FSDP+SP4 & 64k  & 79.10  & OOM     & \textbf{468}   & -    &  43.98        & -  \\
FSDP+SP8 & 64k  & 70.09  & 79.04   & \textbf{441}   & 335  &  41.63        & 31.56\\
FSDP+SP4 & 128k & OOM    & OOM     & -              & -    &  -            & -\\
FSDP+SP8 & 128k & 76.67  & OOM     & \textbf{303}   & -    &  43.68        & -\\
\bottomrule
\end{tabular}
\end{table}

\begin{table}[H]
\centering
\caption{\textbf{Performance of \veomni with Qwen3-30B-A3B~\cite{qwen3technicalreport} on 128GPUs.} We adopt mixed-precision training and full activation checkpointing, with a micro-batch size of 1 and a global batch size of 128. (TorchTitan does not support this setting.)}
\label{tab: qwen3moe-30b table}
\begin{tabular}{lrrrrr}
\toprule
\textbf{Method}   & \textbf{Seqlen} &  \textbf{Memory (GB)}  & \textbf{Throughput} &\textbf{MFU (\%)}\\
\midrule
FSDP+SP1+EP8          & 8k                      &   45.58          & 2,776               & 7.45     \\
FSDP+SP4+EP8          & 8k                      &   22.56          & 1,216               & 3.02     \\
FSDP+SP1+EP8          & 16k                     &   69.99          & 2,853               & 10.74    \\
FSDP+SP4+EP8          & 16k                     &   34.32          & 1,897               & 6.55     \\
FSDP+SP1+EP8          & 32k                     &   77.25          & 2,417               & 13.40    \\
FSDP+SP4+EP8          & 32k                     &   63.08          & 2,064               & 11.35    \\
FSDP+SP1+EP8          & 64k                     &   OOM            & -                   & -      \\
FSDP+SP4+EP8          & 64k                     &   77.92          & 1,716               & 15.65    \\
FSDP+SP1+EP8          & 128k                    &   OOM            & -                   & -      \\
FSDP+SP4+EP8          & 128k                    &   78.01          & 1,075               & 17.92    \\
\bottomrule
\end{tabular}
\end{table}

These benchmark results demonstrate that \veomni consistently achieves higher throughput and memory efficiency across a range of model sizes and sequence lengths. Notably, \veomni supports long-sequence and MoE model training that exceed the memory or capability limits of TorchTitan.

\clearpage

\section{\veomni's API Design}
\subsection{Data Processing and Collating}
\label{appendix:omni-data}

Listing~\ref{lst:veomni-data-collator} demonstrates the data collation pipeline in \veomni, which provides a unified interface for processing and batching omni-modal inputs, including text, image, video, and audio. The \texttt{OmniDataCollatorWithPacking} class enables fine-grained control over feature packing and concatenation, making it straightforward to prepare batched training data for diverse modalities. This modular design simplifies the integration of new modalities and supports flexible token-level and feature-level alignment strategies. The resulting \texttt{train\_dataloader} is constructed using \texttt{build\_dataloader()}, which automatically aligns with the parallel training configuration.

\begin{lstlisting}[language=Python, caption={Data Processing and Collating}, label={lst:veomni-data-collator}]
from veomni.data import (
    OmniDataCollatorWithPacking, build_dataloader
)

data_collate_fn.append(
    OmniDataCollatorWithPacking(
        packing_features=[
            "input_ids",
            "attention_mask",
            "labels",
            "position_ids",
            "image_input_mask",
            "image_output_mask",
            "video_input_mask",
            "video_output_mask",
            "audio_input_mask",
            "audio_output_mask",
        ],
        concat_features=[
            "image_input_features",
            "image_output_features",
            "video_input_features",
            "video_output_features",
            "audio_input_features",
            "audio_output_features",
        ],
    )
)

train_dataloader = build_dataloader(
    dataset=train_dataset,  
    micro_batch_size=micro_batch_size,
    global_batch_size=global_batch_size,
    collate_fn=data_collate_fn,  
)
\end{lstlisting}

\subsection{Omni-Modal LLMs Initialization}
\label{appendix:Omni-model}

Listing~\ref{lst:veomni-build-omni-model} shows the usage of \veomni's \texttt{build\_omni\_model} API, which serves as the unified entry point for constructing omni-modal models. Given the configuration file and model component paths (including modality-specific encoders, decoders, and the foundation model), this function automatically assembles and initializes the model on a designated device. By abstracting away the underlying initialization logic, \veomni ensures that users can easily instantiate complex omni-modal models in a modular and scalable fashion, while maintaining compatibility with distributed parallelization workflows.

\begin{lstlisting}[language=Python, caption={Omni-Modal LLMs Initialization}, label={lst:veomni-build-omni-model}]
from veomni.models import build_omni_model

model = build_omni_model(
    config_path=config_path,
    weights_path=model_path,
    encoders=encoders,
    decoders=decoders,
    foundation=foundation,
    init_device=init_device,
)
\end{lstlisting}

\subsection{Omni-Modal Model Computation}
\label{appendix:Omni-model_train-infer}

Listing~\ref{lst:veomni-train-omni-model}-\ref{lst:veomni-infer-omni-model} illustrates the partial forward process implementation of the plug-and-play architecture in \veomni, named OmniModel, which adopts an encoder–foundation–decoder architecture. The snippet demonstrates how the forward logic across training and inference modes integrates each module (\ie, encoder, foundation model, and decoder) in a seamless and modular pipeline. During inference, the control flow dynamically switches to modality-specific generation branches based on the generated text special tokens, enabling flexible omni-modal outputs. During inference, OmniModel supports parallelized generation and classifier-free guidance~\cite{wu2025janus}. This architecture highlights the clear decoupling between encoder, foundation, and decoder modules, allowing each component to define and execute its own forward logic independently.

\begin{lstlisting}[language=Python, caption={OmniModel Training Process}, label={lst:veomni-train-omni-model}]
class OmniEncoder:
    def forward(**inputs):
        inputs_embeds = self.text_encoder(inputs['input_ids'])
        image_input_embeds = self.image_encoder.lm_encode(
            inputs.get("image_input_features")
        )
        video_input_embeds = self.video_encoder.lm_encode(
            inputs.get("video_input_features")
        )
        audio_input_embeds = self.audio_encoder.lm_encode(
            inputs.get("audio_input_features")
        )
        inputs_embeds = self.masked_scatter(
            inputs_embeds, 
            image_input_embeds,
            inputs.get("image_input_mask"),
            video_input_embeds,
            inputs.get("video_input_mask"),
            audio_input_embeds,
            inputs.get("audio_input_mask"),
        )
        return inputs_embeds

class OmniDecoder:
    def forward(inputs_embeds, **inputs):
        image_output_embeds = self.image_decoder.lm_encode(
            inputs.get("image_output_features")
            )
        video_output_embeds = self.video_encoder.lm_encode(
            inputs.get("video_output_features")
        )
        audio_output_embeds = self.audio_encoder.lm_encode(
            inputs.get("audio_output_features")
        )
        inputs_embeds = self.masked_scatter(
            inputs_embeds, 
            image_output_embeds,
            inputs.get("image_output_mask"),
            video_output_embeds,
            inputs.get("video_output_mask"),
            audio_output_embeds,
            inputs.get("audio_output_mask"),
        )
        return inputs_embeds

    def lm_head(hidden_states, **inputs):
        loss = self.image_decoder.lm_head(outputs.hidden_states, **inputs)
        loss += self.video_decoder.lm_head(outputs.hidden_states, **inputs)
        loss += self.audio_decoder.lm_head(outputs.hidden_states, **inputs)
        return loss
        

class OmniModel:
    encoder: OmniEncoder
    decoder: OmniDecoder
    
    def forward(self, **inputs):
        inputs_embeds = self.encoder(**inputs)
        inputs_embeds = self.decoder(inputs_embeds, **inputs)
        outputs = self.foundation(inputs_embeds, **inputs)

        hidden_states = outputs.hidden_states
        loss = None
        
        if self.training:
            loss = outputs.loss
            loss += self.decoder.lm_head(hidden_states, **inputs)

        return OmniOutput(
            loss = loss,
            hidden_states = hidden_states
        )
        
\end{lstlisting}

\begin{lstlisting}[language=Python, caption={OmniModel Inference Process}, label={lst:veomni-infer-omni-model}]
class OmniModel:
    def setup_image_generation():
        self.generation_type = "image"
        self.setup_parallelize_generation()
        self.setup_classifier_free_guidance()
        self.setup_position_id_map()

    def setup_video_generation():
        ...

    def setup_audio_generation():   
        ...

    def setup_text_generation():
        ...
            
    def prepare_inputs_for_generation(self, **inputs):
        model_inputs = self.foundation.prepare_inputs_for_generation(
            **inputs
        )
        if cache_position[0] == 0:
            inputs_embeds = self.encoder(**inputs)
            model_inputs["inputs_embeds"] = inputs_embeds
            return model_inputs

        if self.generation_type == "text":
            if input_ids[0][-1] == self.image_start_token
                self.setup_image_generation()
            if input_ids[0][-1] == self.video_start_token
                self.setup_video_generation()
            if input_ids[0][-1] == self.audio_start_token
                self.setup_audio_generation()

        if self.generation_type == "image":
            hidden_states = model_inputs["hidden_states"]
            input_embeds, next_tokens = self.decoder.image_decoder.lm_embed(
                hidden_states, **self.image_generation_configs
            )
            model_inputs["inputs_embeds"] = input_embeds
            self.image_tokens.append(next_tokens)
            if len(self.image_tokens) == self.image_token_num:
                self.setup_text_generation()
                self.images.append(self.decoder.image_decoder.lm_generate(
                    self.image_tokens
                )
        if self.generation_type == "video":
            ...
        if self.generation_type == "audio":
            ...
            
        return model_inputs
        
\end{lstlisting}

\subsection{Parallel State Management}
\label{appendix:parallel_state}

Listing~\ref{lst:veomni-parallel-state} demonstrates the core interface of \veomni's \texttt{parallel\_state} API, which provides a unified and declarative initialization of all n-D parallelism configurations. Unlike traditional methods~\cite{shoeybi2019megatron} managing parallel states based on process groups, \veomni leverages a DeviceMesh~\cite{liang2025torchtitan} abstraction to represent and organize various parallel topologies, including data parallelism (DP), tensor parallelism (TP), expert parallelism (EP), pipeline parallelism (PP), and sequence parallelism (SP). Once initialized via \texttt{init\_parallel\_state}, users can conveniently manage process groups, ranks, and communication topologies through the global \texttt{parallel\_state} object. This design simplifies the orchestration of n-D parallelism strategies and decouples low-level distributed logic from model implementation.

\begin{lstlisting}[language=Python, caption={Parallel State Management}, label={lst:veomni-parallel-state}]
from veomni.distributed.parallel_state import (
    get_parallel_state, init_parallel_state
)

init_parallel_state(
    dp_size=data_parallel_size,
    dp_replicate_size=data_parallel_replicate_size,
    dp_shard_size=data_parallel_shard_size,
    tp_size=tensor_parallel_size,
    ep_size=expert_parallel_size,
    pp_size=pipeline_parallel_size,
    ulysses_size=ulysses_parallel_size,
    dp_mode=data_parallel_mode,
)

# Get global parallel_state
parallel_state = get_parallel_state()

# Get DP Mesh
dp_mesh = parallel_state.dp_mesh
# Get DP Group
dp_group = parallel_state.dp_group
# Get DP Shard Rank
dp_shard_rank = parallel_state.dp_shard_rank
# Get Ulysses Group
ulysses_group = parallel_state.ulysses_group
\end{lstlisting}

\subsection{Omni-Modal Model Parallelization}
\label{appendix:FSDP}

Listing~\ref{lst:veomni-build-parallelize-model} illustrates the usage of the \texttt{build\_parallelize\_model} API, which applies fully sharded data parallel (FSDP) and a customized parallel plan to the target model. This interface abstracts away low-level distributed training details, such as parameter sharding, precision configuration, and activation checkpointing, enabling efficient large-scale distributed training with minimal code changes and achieving scalability without compromising usability.

\begin{lstlisting}[language=Python, caption={Omni-Modal Model Parallelization}, label={lst:veomni-build-parallelize-model}]
from veomni.distributed.torch_parallelize import build_parallelize_model

model = build_parallelize_model(
    model,
    weights_path=model_path,
    enable_full_shard=enable_full_shard,
    enable_mixed_precision=enable_mixed_precision,
    enable_gradient_checkpointing=enable_gradient_checkpointing,
    init_device=init_device,
    enable_fsdp_offload=enable_fsdp_offload,
)
\end{lstlisting}

\subsection{Long-Context Attention}
\label{appendix:ulysess}

Listing~\ref{lst:veomni-ulysess-flash-attention-forward} presents the implementation of \texttt{flash\_attention\_forward} in \veomni, which extends HuggingFace's native FlashAttention interface to support Ulysses-style sequence parallelism~\cite{jacobs2023deepspeed}. When sequence parallelism is enabled, the query, key, and value tensors are first transformed from sequence-sharded to head-sharded layouts using all-to-all collective operations. After performing attention computation through the standard \texttt{\_flash\_attention\_forward} function, the output is transformed back from head-sharded to sequence-sharded format. This function maintains modular compatibility with HuggingFace models, requiring no changes to attention logic, and enables seamless integration of sequence parallelism with FlashAttention~\cite{dao2022flashattention}.

\begin{lstlisting}[language=Python, caption={Long-Context Attention}, label={lst:veomni-ulysess-flash-attention-forward}]
from transformers.modeling_flash_attention_utils import _flash_attention_forward
from veomni.distributed.parallel_state import get_parallel_state
from veomni.distributed.sequence_parallel import (
    gather_heads_scatter_seq,
    gather_seq_scatter_heads,
)

def flash_attention_forward(
    module: torch.nn.Module,
    query: torch.Tensor,
    key: torch.Tensor,
    value: torch.Tensor,
    attention_mask: Optional[torch.Tensor],
    **kwargs,
) -> Tuple[torch.Tensor, None]:

    # Ulysses all-to-all
    ulysses_enabled = get_parallel_state().ulysses_enabled
    if ulysses_enabled:
        ulysses_group = get_parallel_state().ulysses_group
        
        query = gather_seq_scatter_heads(
            query, seq_dim=1, head_dim=2, group=ulysses_group
        )
        key = gather_seq_scatter_heads(
            key, seq_dim=1, head_dim=2, group=ulysses_group
        )
        value = gather_seq_scatter_heads(
            value, seq_dim=1, head_dim=2, group=ulysses_group
        )

    attn_output: torch.Tensor = _flash_attention_forward(
        query,
        key,
        value,
        attention_mask,
        **kwargs,
    )

    # Ulysses all-to-all
    if ulysses_enabled:
        ulysses_group = get_parallel_state().ulysses_group

        attn_output = gather_heads_scatter_seq(
            attn_output, seq_dim=1, head_dim=2, group=ulysses_group
        )

    return attn_output, None
\end{lstlisting}

\subsection{Expert Parallelism Implementation}
\label{appendix:ep-parallel-plan}

Listing~\ref{lst:veomni-ep-plan} demonstrates the definition of an expert parallelism (EP) plan using \veomni's \texttt{ParallelPlan} interface. This declarative design allows users to specify which modules (e.g., expert projection layers in an MoE block) should be sharded across expert parallelism groups. The syntax follows a wildcard-matching convention similar to PyTorch module names, and leverages DTensor-compatible placement primitives like \texttt{Shard(0)}. When passed to \texttt{build\_parallelize\_model()}, the plan is automatically applied to partition the target model along expert dimensions. This flexible mechanism not only simplifies expert parallelism but also enables clean integration with other parallelism strategies such as tensor parallelism (TP) and FSDP.

\begin{lstlisting}[language=Python, caption={Expert Parallelism Implementation}, label={lst:veomni-ep-plan}]
from veomni.distributed.parallel_plan import ParallelPlan

def get_parallel_plan():
    ep_plan = {
        "model.layers.*.mlp.experts.gate_proj": Shard(0),
        "model.layers.*.mlp.experts.up_proj": Shard(0),
        "model.layers.*.mlp.experts.down_proj": Shard(0),
    }
    parallel_plan = ParallelPlan(
        ep_plan=ep_plan,
    )
    return parallel_plan
\end{lstlisting}

\end{document}